\documentclass[letterpaper, 10 pt, conference]{ieeeconf}  

\IEEEoverridecommandlockouts                              

\usepackage{graphicx} 
\usepackage{xcolor}
\usepackage{amsmath} 
\usepackage{amssymb}  
\usepackage{cite} 
\usepackage{multirow} 
\usepackage{url}
\title{\LARGE \bf
Dual-Horizon Hybrid Internal Model for Low-Gravity Quadrupedal Jumping with Hardware-in-the-Loop Validation}

\author{Haozhe Xu$^{*}$, Yifei Zhao$^{*}$, Wenhao Feng, Zhipeng Wang, Hongrui Sang\\
Cheng Cheng, Xiuxian Li, Zhen Yin and  Bin He
	\thanks{$^{*}$ Equal Contributions}
}

\begin{document}

	\maketitle
	\thispagestyle{empty}
	\pagestyle{empty}


\begin{abstract}
Locomotion under reduced gravity is commonly realized through jumping, yet continuous pronking in lunar gravity remains challenging due to prolonged flight phases and sparse ground contact. The extended aerial duration increases landing impact sensitivity and makes stable attitude regulation over rough planetary terrain difficult. Existing approaches primarily address single jumps on flat surfaces and lack both continuous-terrain solutions and realistic hardware validation. This work presents a Dual-Horizon Hybrid Internal Model for continuous quadrupedal jumping under lunar gravity using proprioceptive sensing only. Two temporal encoders capture complementary time scales: a short-horizon branch models rapid vertical dynamics with explicit vertical velocity estimation, while a long-horizon branch models horizontal motion trends and center-of-mass height evolution across the jump cycle. The fused representation enables stable and continuous jumping under extended aerial phases characteristic of lunar gravity. To provide hardware-in-the-loop validation, we develop the MATRIX (Mixed-reality Adaptive Testbed for Robotic Integrated eXploration) platform, a digital-twin-driven system that offloads gravity through a pulley-counterweight mechanism and maps Unreal Engine lunar terrain to a motion platform and treadmill in real time. Using MATRIX, we demonstrate continuous jumping of a quadruped robot under lunar-gravity emulation across cratered lunar-like terrain.
\end{abstract}
	
\section{INTRODUCTION}
Recent advances in deep-space missions have renewed interest in lunar surface exploration. Regions such as permanently shadowed craters, steep crater rims, and lava tubes are considered scientifically and economically valuable~\cite{arm2023scientific, valsecchi2023towards}. However, these areas are characterized by irregular terrain, loose regolith, and significant slope variations. Conventional wheeled rovers exhibit limited mobility under such conditions due to slip and sinkage. Legged robots, in contrast, provide discrete footholds and adaptable contact configurations, making them promising candidates for future planetary exploration tasks in complex environments.

Under lunar gravity, locomotion characteristics change substantially~\cite{lunarleap2024}. Reduced gravitational acceleration leads to prolonged flight phases, sparse ground contact, and increased sensitivity to landing impact. Dynamic gaits such as jumping or pronking are therefore more energy-efficient than continuous walking~\cite{kolvenbach2018efficient}, yet achieving stable and continuous jumping over rough lunar terrain requires coordinated mid-air attitude regulation, impact-aware landing control, and terrain adaptation. Several studies have addressed low-gravity jumping. Kolvenbach~\textit{et~al.}~\cite{kolvenbach2019spacebok} introduced the SpaceBok quadruped designed specifically for lunar exploration, demonstrating energy-efficient repetitive jumping enabled by parallel elastic leg actuation and reaction-wheel-based attitude stabilization under simulated lunar gravity. Rudin~\textit{et~al.}~\cite{rudin2021catlike} further explored deep reinforcement learning for low-gravity locomotion, enabling cat-like mid-air reorientation and controlled landing through non-holonomic limb coordination. The Lunar Leap Robot~\cite{lunarleap2024} adopts a hybrid reinforcement learning architecture with multitask experience sharing to learn obstacle-crossing jumps under lunar gravity. However, these paradigms primarily address isolated jumps or aerial attitude control~\cite{qi2023reinforcement, qi2024reinforcement}, while continuous sequential pronking across three-dimensional rough terrain remains largely unexplored.

\begin{figure}[t]
	\centering
	\includegraphics[width=1\linewidth]{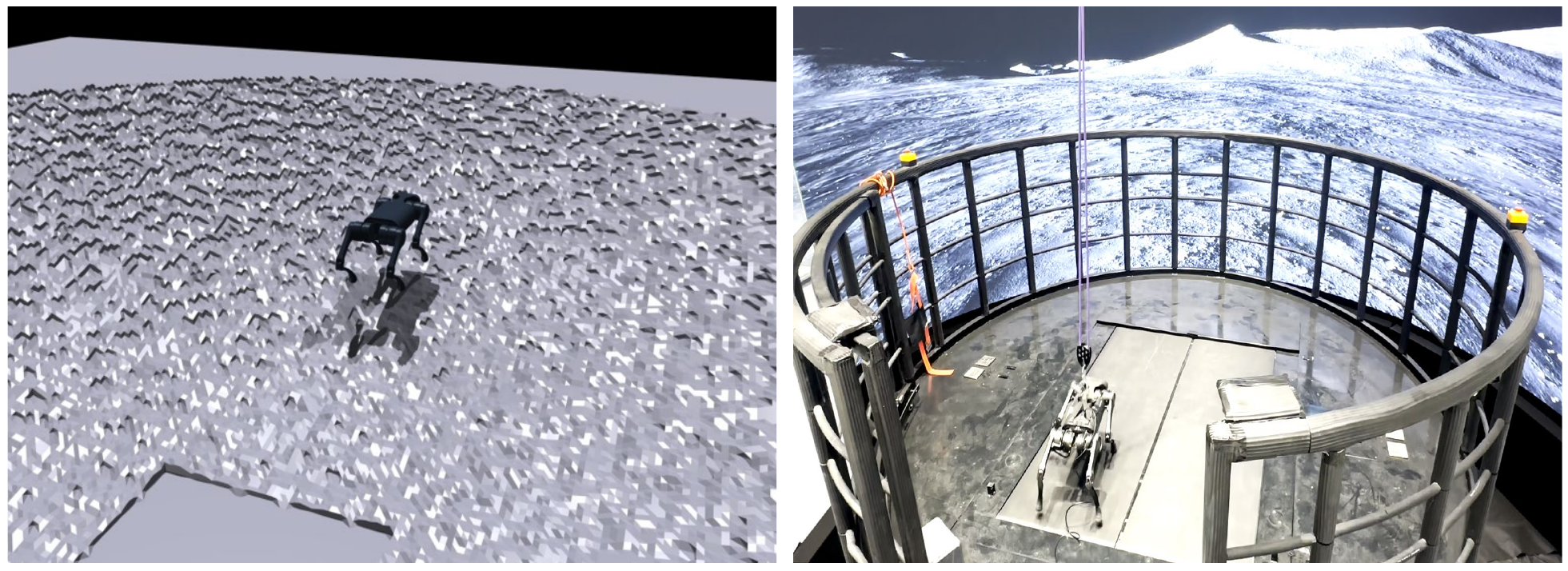} 
	\caption{\textbf{Left:} Training the quadruped robot in the Isaac Gym simulator under lunar gravity. \textbf{Right:} Deploying the learned policy on our MATRIX system while simultaneously emulating low gravity and lunar terrain. }
	\label{fig:intro}
\end{figure}

Achieving continuous jumping demands robust state estimation throughout the prolonged aerial phase. Under Earth gravity, early studies on quadrupedal jumping primarily relied on model-based trajectory optimization and model predictive control~\cite{nguyen2019optimized, park2021jumping, song2022optimal}. Although these approaches provide rigorous dynamic feasibility, they depend heavily on accurate system models and exhibit limited scalability to adaptive or consecutive jumps over irregular terrain. In recent years, deep reinforcement learning has shown promising results in learning agile quadrupedal locomotion behaviors~\cite{Zhuang2023RobotPL, Cheng2023ExtremePW, zhang2024learning}. Hierarchical learning-control frameworks have been introduced to combine reinforcement learning with model-based controllers, enabling continuous and adaptive jumping behaviors by generating high-level motion commands that are tracked by low-level optimization-based leg controllers~\cite{yang2023cajun}. In~\cite{atanassov2024curriculum}, a curriculum-based reinforcement learning framework is proposed to facilitate the learning of dynamic jumping skills without relying on reference trajectories, where progressively structured training tasks guide the policy from simple vertical jumps to long-distance and obstacle-crossing maneuvers. More recent studies extend learning-based jumping toward omnidirectional and height-aware behaviors by jointly training locomotion policies with auxiliary state estimation modules that infer body and foot states during aerial motion~\cite{han2025omninet}. Methods such as Hybrid Internal Model~\cite{long2024hybrid} and DreamWaQ~\cite{nahrendra2023dreamwaq} encode short proprioceptive histories to infer implicit terrain and state information. However, under lunar gravity the aerial phase is significantly prolonged and contact events become sparse and discontinuous, so that a single short observation window cannot capture the full state evolution across a complete jump cycle, limiting the effectiveness of short-horizon state reconstruction. To address this, we propose a Dual-Horizon Hybrid Internal Model employing two temporal encoders at complementary time scales, together with a Phase-Adaptive Gated Reward for stage-dependent regulation of takeoff, flight, and landing.

Validating low-gravity locomotion on physical hardware is equally challenging. Air-bearing facilities approximate reduced gravity in constrained planar environments~\cite{rudin2021catlike, qi2024reinforcement}, while suspension-based systems offset gravity via cables or springs. SpaceHopper~\cite{Spiridonov2024SpaceHopperAS} demonstrates vertical jumping via a pulley--counterweight setup and two-axis attitude reorientation in a gimbal. Arm~\textit{et~al.}~\cite{arm2025efficient} employ a constant-force spring offload system for lunar-gravity walking experiments with Magnecko, yet the tests are limited to flat-ground locomotion without terrain variation. More broadly, all existing setups operate over static flat terrain and validate only a single locomotion mode rather than continuous multi-jump trajectories over three-dimensional cratered surfaces. To close this gap, we develop the MATRIX (Mixed-reality Adaptive Testbed for Robotic Integrated eXploration) platform, a digital-twin-driven hardware-in-the-loop system that emulates lunar gravity through a pulley-counterweight mechanism and reproduces lunar terrain in real time by mapping an Unreal Engine virtual environment to a motion platform and treadmill.

The main contributions of this work are summarized as follows:

		\begin{itemize}
	
		\item A Dual-Horizon control framework for continuous quadrupedal jumping over complex lunar terrain, incorporating a multi-timescale internal model and a phase-adaptive reward design.
		\item The development of the MATRIX hardware-in-the-loop digital twin platform, which enables real-time emulation of lunar gravity and three-dimensional terrain geometry.
		\item Experimental demonstration of continuous quadrupedal jumping under lunar-gravity emulation across cratered lunar-like terrain using the proposed platform.
		
	\end{itemize}

\begin{figure*}[t]
    \centering
    \includegraphics[width=\linewidth]{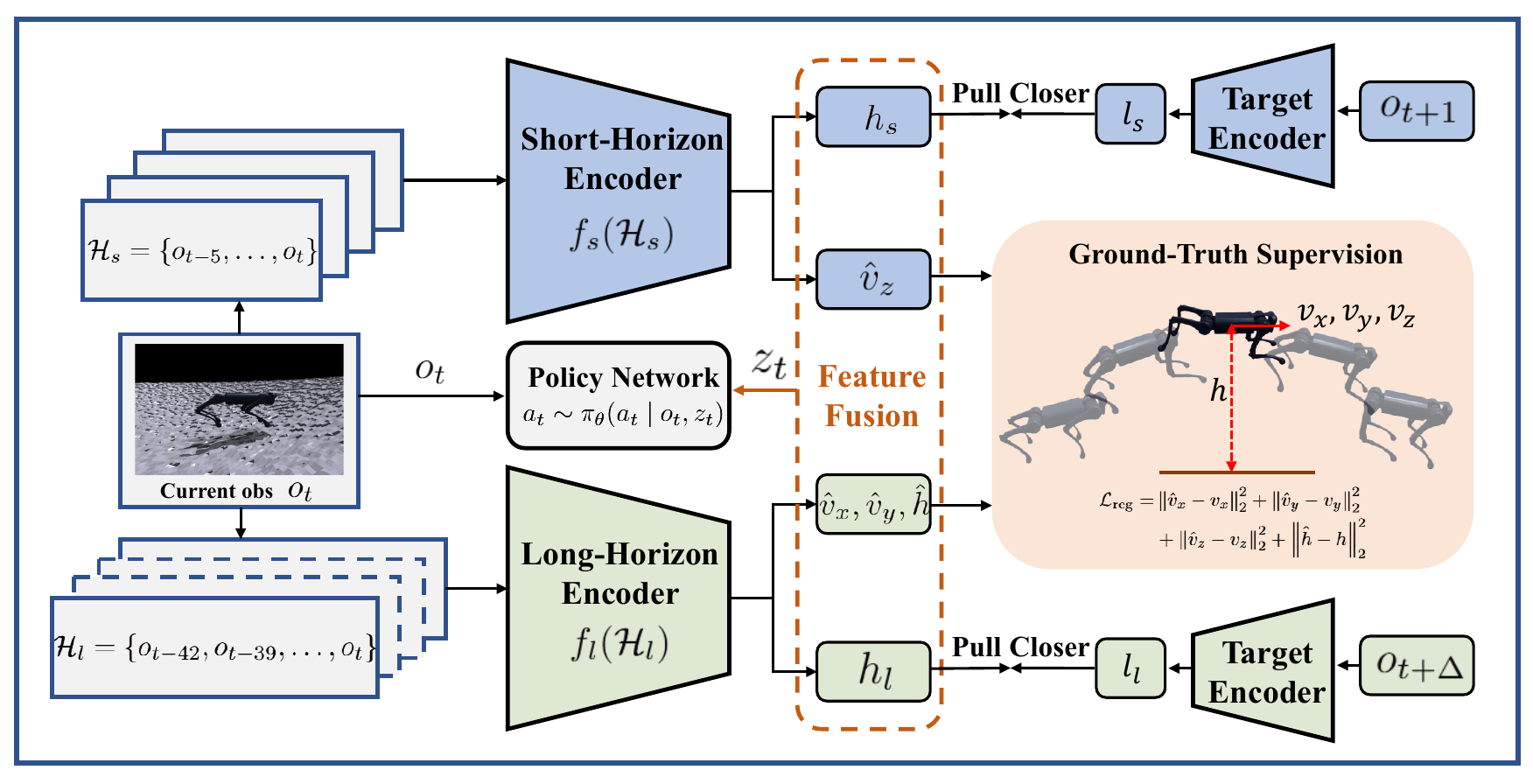}
    \caption{Overview of the proposed Dual-Horizon Hybrid Internal Model. Two temporal encoders process proprioceptive history at complementary time scales. The short-horizon encoder $f_s$ operates on the recent window $\mathcal{H}_s$ (6 steps, $\approx$0.12\,s) to capture rapid vertical dynamics, producing an explicit vertical-velocity estimate $\hat{v}_z$ for phase awareness and a latent vector $h_s$. The long-horizon encoder $f_l$ operates on a subsampled window $\mathcal{H}_l$ ($\approx$0.9\,s) to track slow trends across the full jump cycle, producing horizontal-velocity estimates $\hat{v}_x, \hat{v}_y$, CoM-height estimate $\hat{h}$, and a latent vector $h_l$. The fused representation $z_t = [\hat{v}_x, \hat{v}_y, \hat{v}_z, \hat{h}, h_s, h_l]$ augments the current observation $o_t$ as input to the policy network. Training combines ground-truth regression supervision ($\mathcal{L}_{\mathrm{reg}}$) for the explicit estimates with InfoNCE-style contrastive objectives for the latent vectors, where each branch is pulled toward a target-encoder embedding of a future observation.}
    \label{fig:overview}
\end{figure*}

\section{Dual-Horizon Hybrid Internal Model for Low-Gravity Jumping}

\subsection{Overview}

The proposed framework targets stable, continuous quadrupedal jumping over complex terrain under lunar gravity using proprioceptive sensing only. The task is formulated as a Markov Decision Process and optimized with Proximal Policy Optimization~(PPO), which constrains each update within a trust region for stable training. The Dual-Horizon Hybrid Internal Model (Sec.~\ref{sec:dhhim}) employs two temporal encoders: a short-horizon branch (Sec.~\ref{sec:temporal}) explicitly estimates vertical velocity for phase awareness, while a long-horizon branch captures horizontal motion trends and CoM height evolution across the full jump cycle, with their fused representation augmenting the policy input without external terrain sensing. A Phase-Adaptive Gated Reward (Sec.~\ref{sec:reward}) activates stage-dependent objectives inferred from physically interpretable quantities to regulate takeoff, flight, and landing. The training terrain curriculum is described in Sec.~\ref{sec:terrain}. An overview of the system is shown in Fig.~\ref{fig:overview}.

\subsection{Dual-Horizon Hybrid Internal Model}
\label{sec:dhhim}

Hybrid Internal Model encodes a short history of proprioceptive observations using an estimator network, which outputs an explicit velocity estimate and a latent vector representing implicit environmental response. This mechanism enables robust locomotion without external terrain sensing.

The original formulation employs a short temporal window of six time steps, corresponding to approximately $0.12$ seconds at a control frequency of $50$ Hz. This time scale is sufficient for ground locomotion under Earth gravity.

In low-gravity continuous jumping, the system dynamics evolve over significantly longer time scales. For a quadruped robot, a jump that raises the center of mass by approximately $0.3\,\mathrm{m}$ 
results in an aerial duration of about $1.2\,\mathrm{s}$ under lunar gravity. 
Under Earth gravity, the same height change yields an aerial duration of about $0.5\,\mathrm{s}$. 

Therefore, the flight phase under lunar gravity is more than twice as long as the Earth-gravity equivalent for the same jump height. Over such an extended aerial phase, ground contact events are sparse and body-state evolution is slow relative to the control frequency, so a short observation window covers only a small fraction of the jump cycle and cannot encode the state context needed for stable takeoff, flight, and landing.

For this reason, multi-timescale response modeling is introduced through a dual-horizon internal model.

\subsubsection{Observation Space}

The proprioceptive observation at time step $t$ is defined as $o_t = (v^{\text{cmd}}_t, \omega_t, g_t, q_t, \dot{q}_t, a_{t-1})$, where $v^{\text{cmd}}_t$ denotes the commanded linear velocity, $\omega_t \in \mathbb{R}^3$ denotes the base angular velocity in the body frame, $g_t \in \mathbb{R}^3$ denotes the projected gravity vector, $q_t \in \mathbb{R}^{12}$ denotes joint positions, $\dot{q}_t \in \mathbb{R}^{12}$ denotes joint velocities, and $a_{t-1} \in \mathbb{R}^{12}$ denotes the previous action.


\subsubsection{Action Space}
The action $a_t \in \mathbb{R}^{12}$ represents the desired joint position increment relative to a nominal configuration $\bar{q}$, where $a_t = q^*_t - \bar{q}$ and $q^*_t$ denotes the desired joint position. The desired joint position $q^*_t$ is tracked by a joint-level proportional-derivative controller.

\subsubsection{Dual-Horizon Temporal Modeling}
\label{sec:temporal}

Two temporal windows are constructed from proprioceptive observations.

\textbf{Short Horizon.}
$\mathcal{H}_s = \{o_{t-5},\dots,o_t\}$ (6 steps, $\approx$0.12\,s) is encoded by $f_s$ into $z_s = [\hat{v}_z,\, h_s]$, providing an explicit vertical-velocity estimate $\hat{v}_z$ for phase awareness and a short-horizon latent $h_s$.

\textbf{Long Horizon.}
$\mathcal{H}_l = \{o_{t-42}, o_{t-39},\dots, o_t\}$ (15 subsampled frames, $\approx$0.9\,s) is encoded by $f_l$ into $z_l = [\hat{v}_x,\hat{v}_y,\hat{h},\, h_l]$, capturing horizontal motion trends and CoM height $\hat{h}$ across the full jump cycle.

\textbf{Feature Fusion.}
The fused representation $z_t = [\hat{v}_x, \hat{v}_y, \hat{v}_z, \hat{h}, h_s, h_l]$ augments the policy input: $a_t \sim \pi_\theta(a_t \mid o_t, z_t)$.

\textbf{Supervised regression losses.}
The estimator is supervised with ground-truth simulator signals via
\begin{equation}
\mathcal{L}_{\mathrm{reg}} = \sum_{q\in\{v_x,v_y,v_z,h\}} \|\hat{q} - q\|_2^2.
\end{equation}

\textbf{Contrastive objectives.}
Both branches use an InfoNCE-style contrastive loss. For branch $b \in \{s, l\}$ with temporal offset $\delta_b$ ($\delta_s=1$, $\delta_l=\Delta$):
\begin{equation}
\mathcal{L}_{b} =
-\log
\frac{
\exp\left(\mathrm{sim}\big(h_b(t),\, \phi(o_{t+\delta_b})\big) / \tau \right)
}{
\sum_{j \in \mathcal{B}}
\exp\left(\mathrm{sim}\big(h_b(t),\, \phi(o_{j})\big) / \tau \right)
},
\end{equation}
where $\phi(\cdot)$ is a target encoder, $\tau$ the temperature, $\mathcal{B}$ the minibatch, and $o_{j}$ denotes the proprioceptive input of sample $j$. The offset $\Delta=6$ (0.12\,s) is chosen to match the near-ground contact phase, allowing $h_l$ to capture slow jump-cycle trends while reducing contact-event noise.

\textbf{Overall loss.}
The estimator is trained by minimizing the weighted sum of the above objectives:
\begin{equation}
\mathcal{L}_{\mathrm{DH}} =
\lambda_{\mathrm{reg}} \mathcal{L}_{\mathrm{reg}}
+
\lambda_{s} \mathcal{L}_{s}
+
\lambda_{l} \mathcal{L}_{l},
\end{equation}
where $\lambda_{\mathrm{reg}}$, $\lambda_{s}$, $\lambda_{l}$ are scalar weights.

\subsection{Phase-Adaptive Gated Reward Design}
\label{sec:reward}

Rather than using contact sensors or explicit phase variables, the proposed reward infers the jumping stage from CoM height and vertical velocity, activating stage-specific objectives without a state machine.

The total reward at time step $t$ is defined as
\begin{equation}
\begin{aligned}
R_t =\;& R_t^{global}
+ w_t^{takeoff} R_t^{takeoff} \\
&+ w_t^{flight} R_t^{flight}
+ w_t^{land} R_t^{land}.
\end{aligned}
\end{equation}

\subsubsection{Phase Activation Factors}

Three binary indicators are derived from CoM height $h_t$ and vertical velocity $v_{z,t}$: a near-ground flag $G_t = \mathbf{1}[h_t \le h_{thr}]$, an upward flag $I_t^{up} = \mathbf{1}[v_{z,t} > v_{thr}]$, and a downward flag $I_t^{down} = \mathbf{1}[v_{z,t} < -v_{thr}]$, where $v_{thr} = \alpha\sqrt{2g_{\mathrm{moon}}(h_{cmd}-h_{stance})}$ with $\alpha=0.5$. The phase weights are then
\begin{equation}
w_t^{\text{takeoff}} = G_t I_t^{up},\quad
w_t^{\text{land}}    = G_t I_t^{down},\quad
w_t^{\text{flight}}  = 1 - G_t,
\label{eq:phase_weights}
\end{equation}
partitioning the motion into takeoff, aerial, and landing regimes without explicit phase scheduling.

\subsubsection{Global Reward Terms}
The global reward applied at every timestep is

\begin{align}
R_t^{\mathrm{global}} 
&= w_v \exp\!\left(-k_v (v_{x,t} - v_x^{\mathrm{cmd}})^2\right)
   - w_r \mathcal{R}_t \nonumber \\
&\quad + R^{\mathrm{peak}} .
\end{align}

The first term tracks the commanded forward velocity $v_x^{cmd}$. 
The second term $\mathcal{R}_t$ denotes a regularization penalty (e.g., action or torque regularization) weighted by $w_r$.

The sparse term $R^{peak} = \mathbf{1}[|\max_t h_t - h_{cmd}| \le \delta_h]$ rewards jumps that reach the commanded height within tolerance $\delta_h = 0.04$\,m, following~\cite{han2025omninet}. The commanded height $h_{cmd}$ is manually specified according to the task requirements.

\subsubsection{Stage-Specific Rewards}
The three stage rewards are
\begin{align}
R_t^{takeoff} &= \exp\!\left(-k_z (v_{z,t} - v_{thr})^2\right), \nonumber\\
R_t^{flight}  &= - w_f (roll_t^2 + pitch_t^2), \\
R_t^{land}    &= - v_{z,t}^2 - w_h \max_i(h_{foot,i}), \nonumber
\end{align}
encouraging sufficient takeoff velocity, mid-air attitude stability, and coordinated touchdown, respectively.

\subsection{Terrain Curriculum Design}
\label{sec:terrain}

\begin{figure}[t]
	\centering
	\includegraphics[width=1\linewidth]{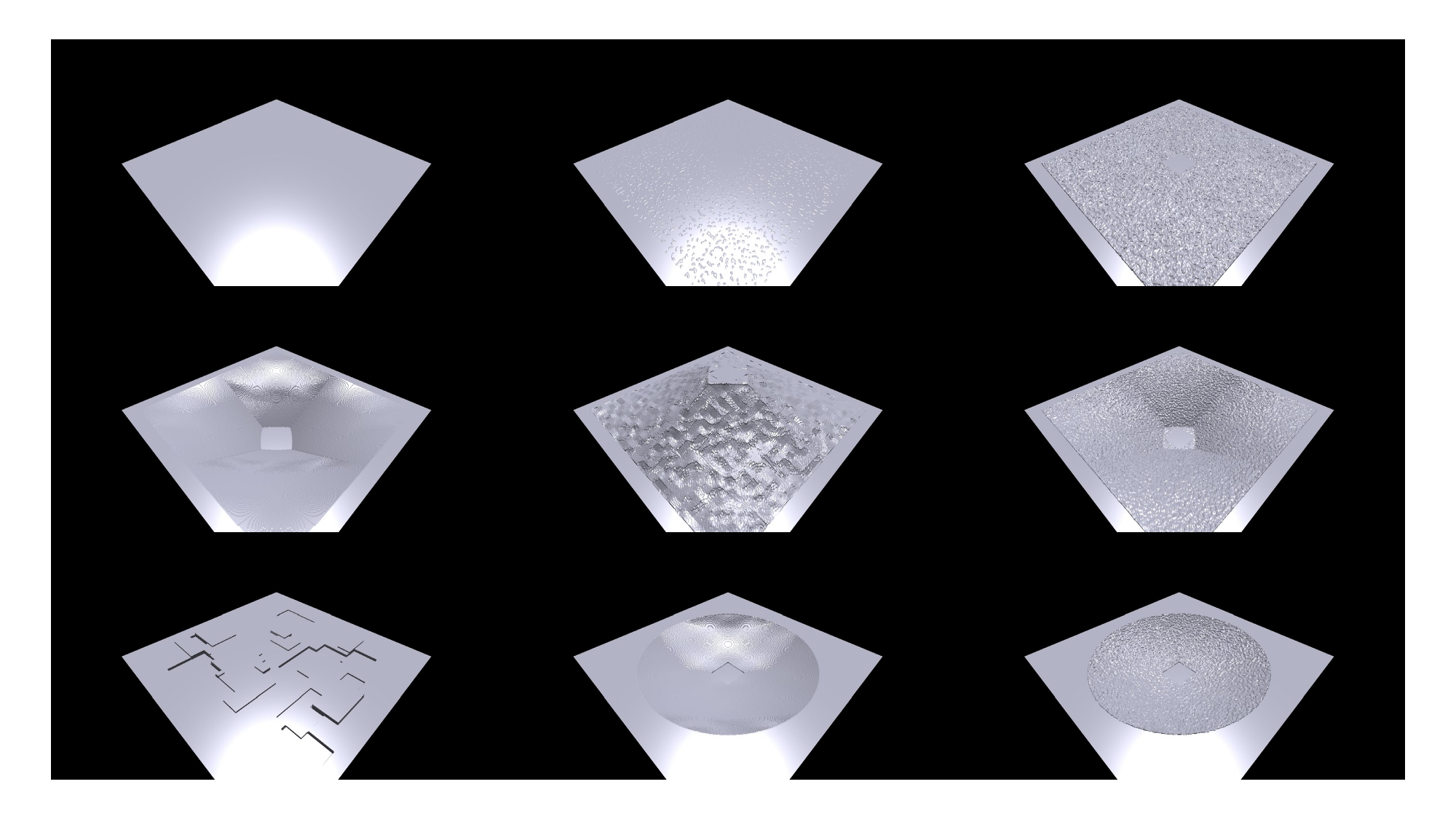} 
	\caption{Nine terrain types used for training. The top row includes Pure Flat, Flat, and Perlin Flat. The middle row includes Smooth Slope, Rough Slope, and Perlin Slope. The bottom row includes Discrete Obstacles, Crater, and Perlin Crater. The Crater and Perlin Crater terrains are defined by an exponential height function to approximate impact crater structures commonly observed on the lunar surface.}
	\label{fig:terrain} 
\end{figure}

To evaluate robustness under diverse surface conditions, a procedurally generated terrain with curriculum-based difficulty scaling is constructed. 
The terrain is organized as a $10 \times 20$ grid of sub-terrain patches, each measuring $8\,\mathrm{m} \times 8\,\mathrm{m}$. 
Rows correspond to monotonically increasing difficulty levels $d \in [0,1]$, while columns are assigned to nine predefined terrain categories according to fixed proportions. All nine terrain categories are illustrated in Fig.~\ref{fig:terrain}. Perlin noise is introduced to further approximate the rough surface textures of the lunar terrain.

\textbf{Surface Roughness Modeling:}
Continuous terrains employ fractal Brownian motion based on two-dimensional Perlin noise. 
At octave $k$, the frequency and amplitude are defined as 
$f_k = f_0 \lambda^k$ and $a_k = z_{\mathrm{scale}} G^k$, 
with $f_0 = 10$, $\lambda = 2.0$, $G = 0.25$, and $K = 2$ octaves. 
The resulting noise fields are summed to generate multi-scale height perturbations.

\textbf{Difficulty Scaling:}
The difficulty parameter $d$ simultaneously controls multiple terrain attributes, including slope inclination, exponential curvature coefficient, obstacle height and density, and Perlin noise amplitude. 
The amplitude is scaled linearly with $d$. 
This multi-parameter scaling ensures gradual and balanced increases in terrain complexity.

\begin{figure*}[t]
    \centering
    \includegraphics[width=\linewidth]{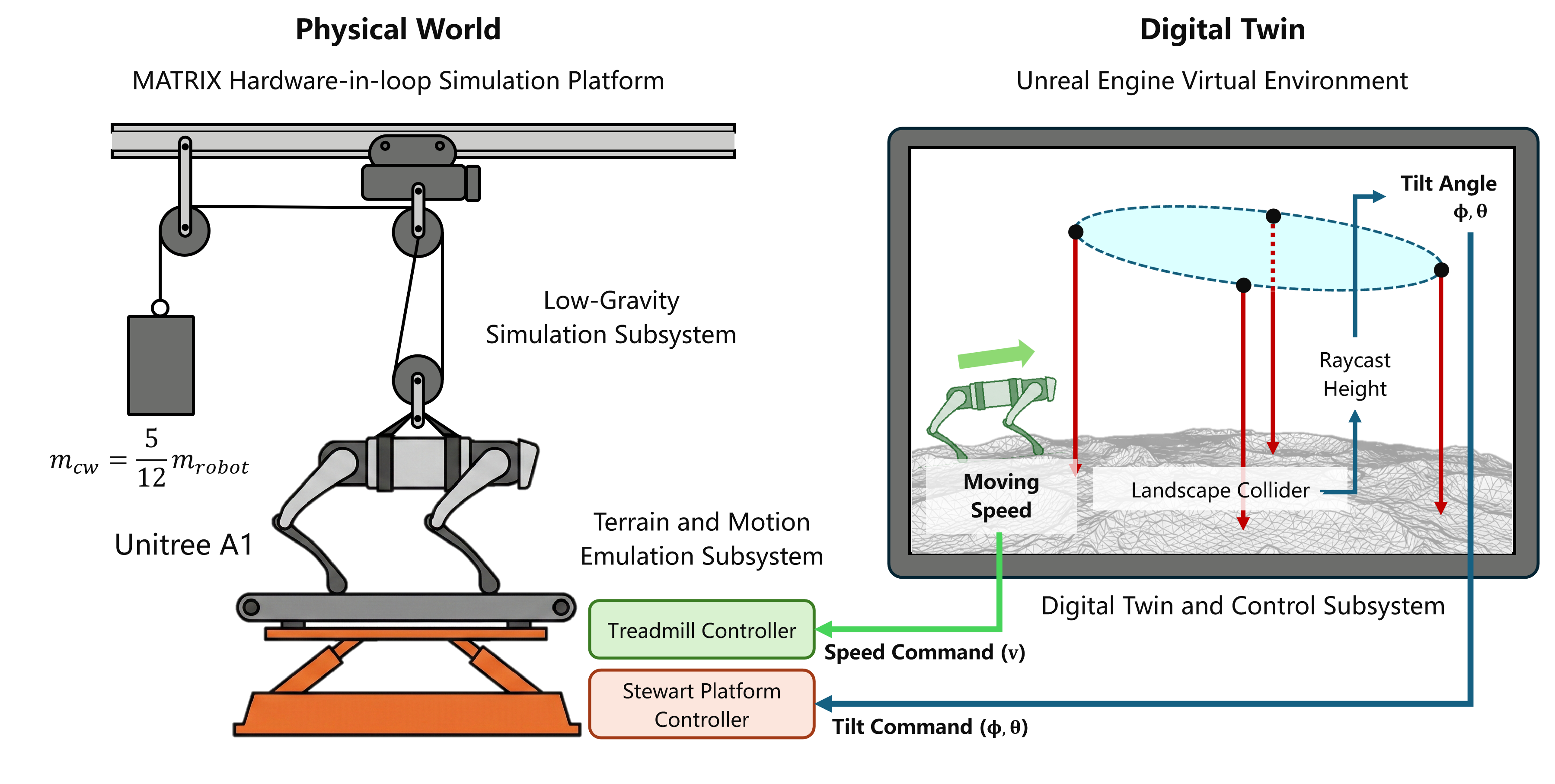}
    \caption{The MATRIX hardware-in-the-loop simulation platform.
             \textit{Left} -- Physical World:
             the \emph{Low-Gravity Simulation Subsystem} uses an overhead
             2:1 pulley block with a counterweight
             ($m_{\mathrm{cw}} = \frac{5}{12}m_{\mathrm{robot}}$)
             to reduce the robot's effective weight to lunar gravity;
             the \emph{Terrain and Motion Emulation Subsystem}
             comprises a 6-DoF Stewart platform and an integrated
             treadmill controlled by the Stewart Platform Controller
             and Treadmill Controller respectively.
             \textit{Right} -- Digital Twin:
             the \emph{Digital Twin and Control Subsystem} runs in
             Unreal Engine; four ray-casts query the Landscape Mesh
             Collider to obtain raycast heights, from which tilt angles
             $(\phi,\theta)$ are computed and sent as a Tilt Command
             to the Stewart Platform Controller, while the virtual robot avatar's
             moving speed is sent as a Speed Command~$v$
             to the Treadmill Controller.}
    \label{fig:matrix_overview}
\end{figure*}

\section{The MATRIX Platform}

To validate the proposed locomotion policy on physical hardware under reduced-gravity and terrain-varied conditions, we develop the \textbf{MATRIX} (\textbf{M}ixed-reality \textbf{A}daptive \textbf{T}estbed for \textbf{R}obotic \textbf{I}ntegrated e\textbf{X}ploration) platform, a digital-twin-driven hardware-in-the-loop testbed that can emulate configurable reduced-gravity conditions and reproduce terrain slope profiles in real time.
As illustrated in Fig.~\ref{fig:matrix_overview}, MATRIX comprises three tightly coupled subsystems:
(i)~a \emph{Low-Gravity Simulation Subsystem} consisting of an overhead pulley-and-counterweight mechanism;
(ii)~a \emph{Terrain and Motion Emulation Subsystem} comprising a six-degree-of-freedom (6-DoF) Stewart-type motion platform with an
integrated motorised treadmill; and
(iii)~a \emph{Digital Twin and Control System} running in Unreal Engine~(UE) that drives subsystems (i) and (ii) through a real-time closed-loop co-simulation pipeline.
In this work, MATRIX is configured to emulate the lunar surface environment, characterised by a gravitational acceleration of
$g_{\mathrm{lunar}} \approx 1.62~\mathrm{m/s^{2}}$ and irregular, cratered regolith terrain. The actual MATRIX platform is shown in Fig.~\ref{fig:intro}.

\textbf{Low-Gravity Simulation Subsystem:}
The \emph{Low-Gravity Simulation Subsystem} offloads the robot's weight to the lunar gravity level using an overhead pulley-and-counterweight mechanism.
A pair of motorised hoists mounted on a ceiling crane rail are connected to the robot's torso harness through a 2:1 mechanical-advantage pulley block.
A static counterweight of mass $m_{\mathrm{cw}}$ is attached to the opposite end of the cable; the pulley doubles the effective lifting force, so the total upward force is $2\,m_{\mathrm{cw}}\,g_{\mathrm{earth}}$.
The counterweight mass is therefore set to $m_{\mathrm{cw}} = \tfrac{5}{12}\,m$, which yields an offload force of $\tfrac{5}{6}\,m\,g_{\mathrm{earth}}$ and reduces the robot's effective weight to $\tfrac{1}{6}\,m\,g_{\mathrm{earth}} = m\,g_{\mathrm{lunar}}$.
The suspension point is spatially fixed; translational locomotion is provided entirely by the treadmill.

\textbf{Terrain Emulation and Digital Twin Control:}
The \emph{Terrain and Motion Emulation Subsystem} and the
\emph{Digital Twin and Control System} operate as a tightly coupled
closed-loop pair, as depicted in Fig.~\ref{fig:matrix_overview}.
The Stewart platform is commanded to tilt in roll~($\phi$) and pitch~($\theta$)
while the remaining four degrees of freedom are held at their neutral positions,
and the integrated treadmill belt runs at a speed matching the robot's forward
velocity so that the robot locomote walk continuously within the finite laboratory workspace.
These two commands are generated in real time by the UE digital twin:
at each simulation tick, four vertical ray-casts are fired downward from
the front~(F), back~(B), left~(L), and right~(R) positions of the robot's
platform footprint, each intersecting the terrain mesh collider and returning
a ground-contact height $h_{\mathrm{F}},\,h_{\mathrm{B}},\,h_{\mathrm{L}},\,h_{\mathrm{R}}$.
A plane fitted to these four points yields unit normal
$\boldsymbol{n} = (n_x,\, n_y,\, n_z)^{\!\top}$.
Adopting an extrinsic $X$-$Y$ rotation convention,
the platform tilt angles are obtained as
\begin{align}
\phi   &= \operatorname{atan2}(n_y,\; n_z), \nonumber\\
\theta &= \operatorname{atan2}\!\bigl(-n_x,\; \sqrt{n_y^{2} + n_z^{2}}\,\bigr),
\label{eq:compute_angle}
\end{align}
where $\phi$ (roll) corresponds to a rotation about the forward axis $x$ and
$\theta$ (pitch) about the lateral axis $y$.
The resulting $(\phi,\,\theta)$ is forwarded to the motion platform
controller, while the robot's virtual forward velocity is sent simultaneously
to the treadmill controller.

\textbf{Digital Twin Terrain Integration in Unreal Engine:}
To construct realistic extra-terrestrial test scenarios, a lunar environment is developed in Unreal Engine 5.6.1 using the \emph{Brushify -- Moon}\cite{brushify_moon_pack} asset package. 
This package provides high-fidelity terrain meshes including irregular regolith surfaces, scattered rocks and pebbles, and large-scale crater formations.

Representative lunar scenes considered in this work are illustrated in Fig.~\ref{fig:ue5brushify}. The terrain distribution and geometric difficulty of the imported lunar level are systematically modified to align with the terrain curriculum described in Section~II-D. 
Specifically, slope inclination, surface roughness amplitude, and crater curvature are adjusted to ensure consistency between the simulation training environments and the physical test scenarios.

Using the plane-fitting and normal-vector-based tilt computation defined in~\eqref{eq:compute_angle}, 
the terrain geometry of the Unreal Engine digital twin is mapped in real time to the \emph{Terrain and Motion Emulation Subsystem}. 
This mapping enables approximate reproduction of the virtual lunar terrain on the physical motion platform, thereby achieving continuous hardware-in-the-loop terrain emulation.

\section{EXPERIMENT RESULTS}

\subsection{Implementation Details}

The proposed jumping policy is trained in the Isaac Gym simulation environment using the Unitree A1 quadruped robot. Training is conducted with 4096 parallel simulation environments to accelerate data collection. Each training run consists of 6000 episodes. The policy is optimized using Proximal Policy Optimization (PPO) with an asymmetric actor--critic architecture. The actor receives only proprioceptive observations while the critic has access to privileged information available only in simulation.

The proposed method introduces a dual-horizon representation that encodes motion dynamics at different temporal scales. The short-horizon encoder adopts a multilayer perceptron with layer dimensions [270,128,64,17], while the long-horizon encoder uses a similar architecture with dimensions [270,128,64,19]. The latent outputs of the two encoders are concatenated and provided to the policy network for action generation.

\subsection{Sim-to-Real Gap Modeling for Suspension-Based Gravity Offloading}

The pulley-counterweight mechanism used in our hardware experiments introduces dynamics discrepancies relative to ideal lunar gravity. Pulley friction, cable elasticity, and counterweight inertia cause tension fluctuations; lateral robot motion tilts the cable and produces horizontal disturbance forces; and rapid takeoff or landing can momentarily slacken the cable, altering the effective gravity. A tension sensor installed on the suspension cable records fluctuations of approximately 0--1.2\,kg equivalent force for a 12.5\,kg robot, representing a notable fraction of the target lunar-gravity load.

To mitigate the sim-to-real gap caused by the suspension system, two simulation adaptations are introduced.

\begin{figure}[t]
	\centering
	\includegraphics[width=1\linewidth]{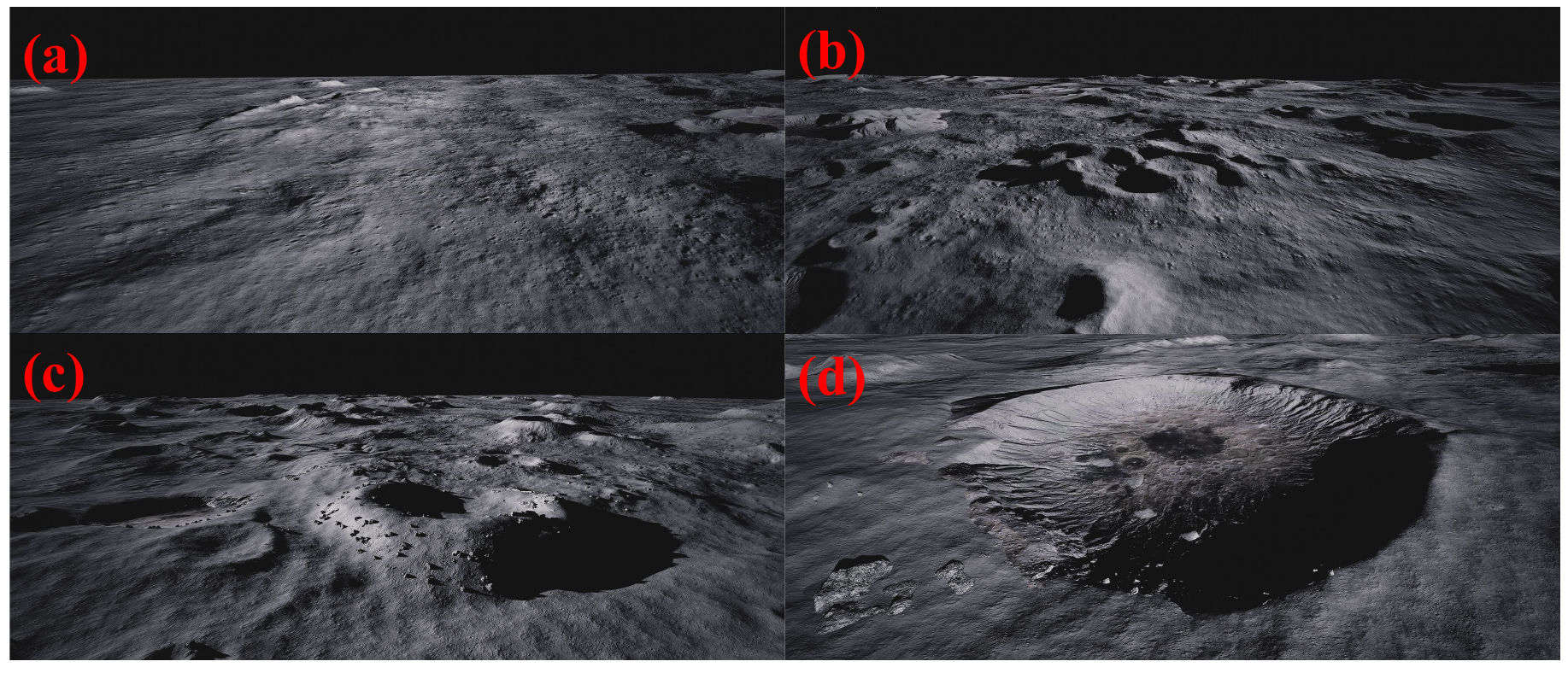} 
	\caption{Representative lunar terrains configured in Unreal Engine. (a) relatively flat mare plains characterized by low surface roughness; (b) irregular uneven ground with multi-scale surface perturbations; (c) undulating hilly terrain with moderate inclination; and (d) crater-like depressions with smooth radial curvature. }
	\label{fig:ue5brushify} 
\end{figure}

\textbf{Gravity Domain Randomization:}
Instead of training the policy under a fixed lunar gravity value, the gravitational acceleration is randomized at the beginning of each episode, where $g \sim \mathcal{U}(g_{\min}, g_{\max})$.
This randomization models variations in effective gravity caused by rope tension fluctuations and mechanical imperfections of the suspension system.

\textbf{Phase-Triggered Disturbance:}
Transient impulse disturbances are injected during takeoff and landing phases. Slack events in the suspension system mainly occur near these transitions. To emulate this phenomenon, short downward force impulses are applied when the robot enters the takeoff or landing phases. These phase transitions are detected using the jump phase indicators defined in~\eqref{eq:phase_weights}. The disturbances simulate sudden changes in effective gravity caused by cable slack.

\subsection{Ablation Studies in Simulation}

We conduct ablation studies in simulation to evaluate the effectiveness of the proposed Dual-Horizon Hybrid Internal Model and the Phase-Adaptive Gated Reward Design. The compared settings are as follows:

\begin{itemize}
    \item \textbf{Dual-Horizon (Ours)}: The complete method with the Dual-Horizon Hybrid Internal Model and the Phase-Adaptive Gated Reward Design.
    
    \item \textbf{Short-only}: A single short-horizon encoder is used to estimate $\hat{v}_x$, $\hat{v}_y$, $\hat{v}_z$, and $\hat{h}$ from the short temporal window only.
    
    \item \textbf{Long-only}: A single long-horizon encoder is used to estimate $\hat{v}_x$, $\hat{v}_y$, $\hat{v}_z$, and $\hat{h}$ from the long temporal window only.
    
    \item \textbf{w/o Phase Reward}: The proposed Dual-Horizon Hybrid Internal Model is retained, while the Phase-Adaptive Gated Reward Design is removed and replaced by a non-phase-specific reward formulation.
\end{itemize}

To quantitatively evaluate the effectiveness of the proposed Dual-Horizon Hybrid Internal Model and the Phase-Adaptive Gated Reward Design, we consider two categories of metrics: state estimation accuracy and jumping control performance.

\begin{figure}[t]
	\centering
	\includegraphics[width=1\linewidth]{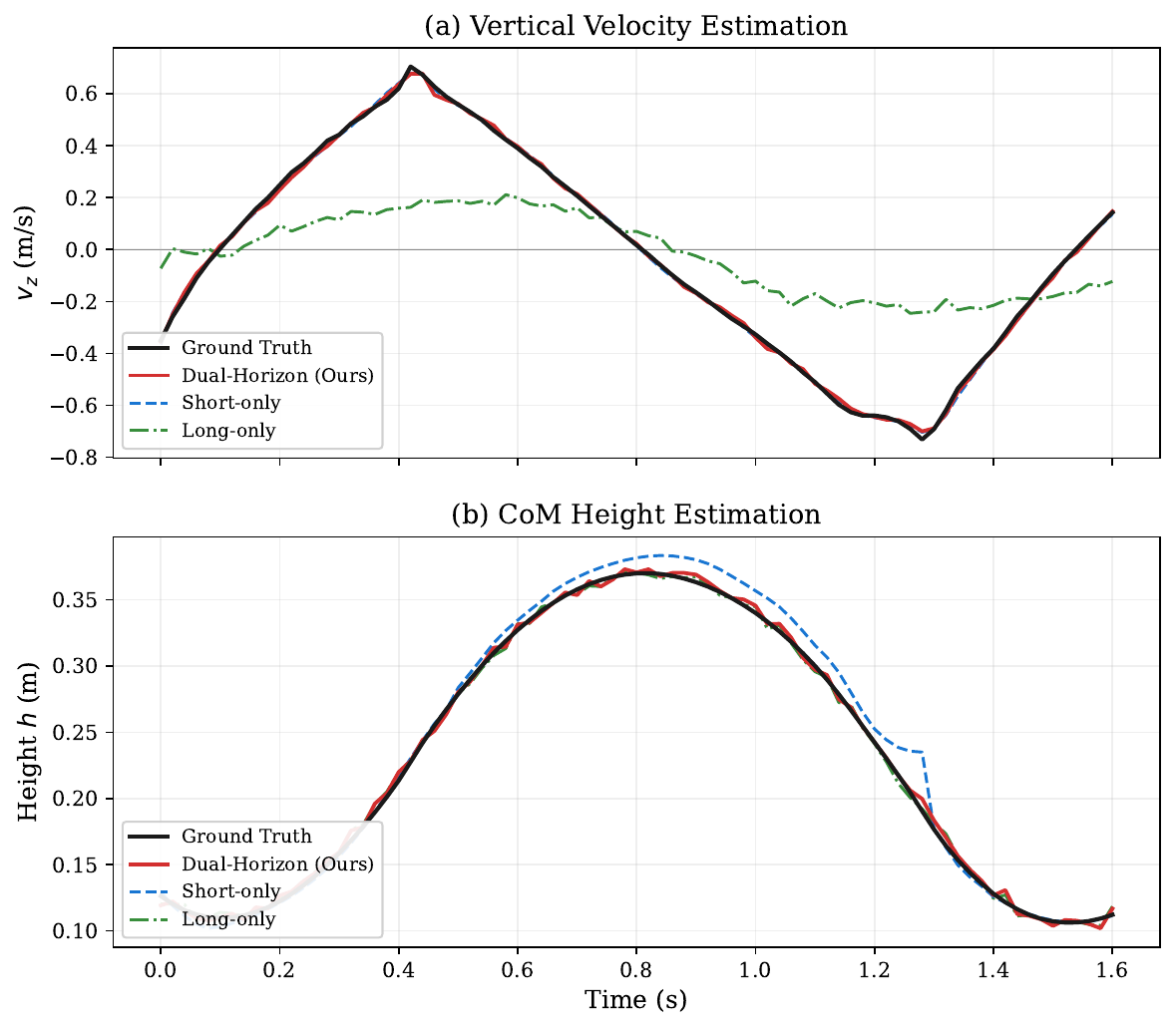} 
	\caption{Comparison of state estimation across different temporal modeling strategies during a representative jump cycle. }
	\label{fig:plot} 
\end{figure}

\begin{table*}[t]
\centering
\caption{Ablation study in simulation.}
\begin{tabular}{c|cccc|ccc}
\hline
Method & $v_z$ MSE & $v_x$ MSE & $v_y$ MSE & $h$ MSE & Survival (s) & Landing SR (\%) & Height Err. (m) \\
\hline
Short-only & 0.041 & 0.013 & 0.013 & 0.028 & 13.6 & 71.2 & 0.054 \\
Long-only & 0.067 & 0.012 & 0.011 & 0.014 & 14.8 & 74.5 & 0.041 \\
w/o Phase Reward & 0.036 & 0.014 & 0.013 & 0.016 & 9.7 & 58.4 & 0.073 \\
\textbf{Ours} & \textbf{0.031} & \textbf{0.010} & \textbf{0.009} & \textbf{0.012} & \textbf{18.9} & \textbf{86.7} & \textbf{0.026} \\
\hline
\end{tabular}
\label{tab:ablation_sim}
\end{table*}

\textbf{State Estimation Metrics:}
The first category measures the accuracy of the internal state estimations produced by the temporal encoders. Specifically, we compute the mean squared error (MSE) between the estimated states and the ground-truth simulator states for vertical velocity, horizontal velocity, and center-of-mass (CoM) height, denoted as $\mathrm{MSE}_{v_z}$, $\mathrm{MSE}_{v_x}$, $\mathrm{MSE}_{v_y}$, and $\mathrm{MSE}_{h}$, respectively. These metrics evaluate how well different temporal modeling strategies capture the underlying motion dynamics. In particular, $\mathrm{MSE}_{v_z}$ reflects the accuracy of vertical dynamic estimation during takeoff and landing transitions, while $\mathrm{MSE}_{h}$ evaluates the ability to capture the long-term evolution of CoM height over the jump cycle.

\textbf{Jumping Performance Metrics:}
The second category evaluates the locomotion performance of the learned policies. We measure the \emph{survival time}, defined as the duration before the robot loses balance or falls during continuous jumping. We also compute the \emph{landing success rate}, defined as the percentage of landings in which the robot reaches ground contact while maintaining near-level body orientation. In this work, a landing is considered successful if the body roll and pitch angles remain within a small tolerance around the upright configuration at touchdown. Therefore, the landing success rate reflects the quality of touchdown control, while the survival time measures long-horizon stability across consecutive jumps. In addition, we measure the \emph{peak height tracking error}, which quantifies the deviation between the achieved maximum CoM height and the commanded jump height. These metrics collectively reflect the stability, controllability, and robustness of the jumping behavior.

\paragraph{Effectiveness of Dual-Horizon Temporal Modeling}

We first analyze the estimation accuracy of different temporal modeling strategies. 
Fig.~\ref{fig:plot} (a) compares the estimated vertical velocity $\hat{v}_z$ with the ground-truth signal during a representative jump cycle. 
The Short-only and Dual-Horizon models closely follow the rapid transitions of the vertical velocity at takeoff and landing, while the Long-only model exhibits noticeable delay and smoothing effects. 
The Short-only and Dual-Horizon models closely follow the rapid transitions of the vertical velocity at takeoff and landing, while the Long-only model exhibits noticeably degraded predictions.
This result indicates that short temporal histories are more suitable for capturing fast vertical dynamics.

Fig.~\ref{fig:plot} (b) shows the estimation of the center-of-mass height. 
In contrast to vertical velocity, the Long-only model provides a more accurate estimate of the height evolution across the jump cycle, while the Short-only model exhibits larger fluctuations due to the limited temporal context. 
By combining both time scales, the Dual-Horizon model achieves the most accurate estimation for both vertical dynamics and long-term motion trends.

\paragraph{Quantitative Comparison}

Table~\ref{tab:ablation_sim} presents the quantitative comparison of all methods. 
The Dual-Horizon model achieves the lowest estimation errors across all state variables, confirming that multi-timescale temporal modeling improves internal state inference under prolonged aerial phases. 
In addition, the proposed method achieves the longest survival time and the highest landing success rate, indicating more stable continuous jumping behavior.

\paragraph{Effectiveness of Phase-Adaptive Reward Design}

Removing the phase-adaptive reward leads to significant performance degradation. 
Although the state estimation accuracy remains comparable, the policy without phase-specific rewards exhibits unstable takeoff impulses and poorly controlled touchdown behaviors. 
As shown in Table~\ref{tab:ablation_sim}, the \textbf{w/o Phase Reward} setting results in shorter survival time, lower landing success rate, and larger height tracking error. 
These results demonstrate that phase-dependent reward regulation is essential for learning stable jumping behaviors under low-gravity conditions.

\subsection{Real-World Experiments}

We validate the proposed controller on a Unitree A1 quadruped robot using the MATRIX suspension testbed. We compare our method with several ablation settings in real-world experiments. The compared configurations are as follows:

\begin{itemize}

\item \textbf{Ours}: The proposed method with both Gravity Domain Randomization and Phase-Triggered Disturbance modeling.

\item \textbf{DR}: The policy trained with Gravity Domain Randomization only, without Phase-Triggered Disturbance modeling.

\item \textbf{PD}: The policy trained with Phase-Triggered Disturbance modeling only, without Gravity Domain Randomization.

\item \textbf{None}: The baseline policy trained without Gravity Domain Randomization and without Phase-Triggered Disturbance modeling.

\end{itemize}
 
Experiments are conducted on a treadmill operating at three speeds: 0.3, 0.5, and 0.7\,m/s. The validation terrain is selected from the Unreal Engine digital twin environment, where a relatively flat mare plains region is used to represent a typical lunar terrain for testing.
The robot performs continuous pronking jumps for a maximum duration of 20\,s. Each experimental setting is repeated 10 times and the averaged results are reported to evaluate robustness. Table \ref{realexp} shows the average survival time of the robot before losing balance or falling under different treadmill speeds. Experimental results show that introducing either Gravity Domain Randomization or Phase-Triggered Disturbance modeling improves the robustness of the system to some extent. The proposed method, which combines both \textbf{DR} and \textbf{PD} modeling, achieves the best performance under all treadmill speeds.

\begin{table}[t]
\centering
\caption{Average survival time (s) under different treadmill speeds. }
\begin{tabular}{c|ccc}
\hline
Method & 0.3 m/s & 0.5 m/s & 0.7 m/s \\
\hline
None & 8.4 & 5.2 & 2.1 \\
DR & 12.7 & 9.3 & 5.4 \\
PD & 11.5 & 8.7 & 4.6 \\
\textbf{Ours} & \textbf{20.0} & \textbf{18.3} & \textbf{15.1} \\
\hline
\end{tabular}
\label{realexp}
\end{table}

\begin{table}[t]
\centering
\caption{Average survival time (s) of the proposed method under different lunar terrain scenarios and treadmill speeds. }
\begin{tabular}{c|ccc}
\hline
Terrain & 0.3 m/s & 0.5 m/s & 0.7 m/s \\
\hline
Mare Plains & 20.0 & 18.3 & 15.1 \\
Uneven Ground & 18.6 & 15.9 & 11.8 \\
Hilly Terrain & 17.4 & 14.2 & 10.1 \\
Crater Terrain & 16.8 & 13.5 & 9.4 \\
\hline
\end{tabular}
\label{terrain_generalization}
\end{table}

The proposed method is further validated on the four representative lunar terrain scenarios shown in Fig.~\ref{fig:ue5brushify}, including relatively flat mare plains, irregular uneven ground, undulating hilly terrain, and crater-like depressions. For each terrain, experiments are conducted at treadmill speeds of 0.3, 0.5, and 0.7\,m/s, and the survival time before the robot loses balance or falls is recorded. The results in Table~\ref{terrain_generalization} indicate that the proposed controller maintains relatively stable continuous jumping across all four terrain scenarios at the low speed of 0.3\,m/s. As the treadmill speed increases, the performance gradually degrades. The worst performance is observed on the crater terrain, which is likely caused by the exponential slope geometry that increases the difficulty of foothold stabilization and makes legged locomotion more prone to slipping.

\section{CONCLUSIONS}
This work presents a Dual-Horizon Hybrid Internal Model designed to enable stable and continuous quadrupedal jumping under low-gravity conditions. The framework introduces two temporal encoders that operate at different time scales, allowing the model to capture both rapid vertical dynamics and long-term motion trends across a complete jump cycle. This multi-timescale representation improves the ability of the control policy to handle the prolonged aerial phases that arise under lunar gravity. To evaluate the proposed approach on physical hardware, the MATRIX platform is developed as a hardware-in-the-loop experimental system. Experimental results demonstrate that the proposed method enables stable and continuous quadrupedal jumping across irregular lunar-like terrain under lunar-gravity emulation.

Despite these encouraging results, several limitations remain. Although the MATRIX platform is able to emulate reduced gravity while reproducing lunar terrain geometry, the suspension cable constrains robot motion within a limited spatial region. Furthermore, during highly dynamic movements, variations in cable tension and direction introduce disturbances that cannot be fully eliminated. Although partial compensation is incorporated in simulation, these disturbances still influence the outcomes of physical experiments. In addition, the current platform reproduces only geometric variations of terrain and does not model the contact mechanics between the robot and lunar regolith. Accurate modeling of robot–regolith interaction dynamics remains an important direction for future research in planetary locomotion.

	


%

%
%
%
%
%

\bibliographystyle{IEEEtran}
\bibliography{output.bib}


\end{document}